\documentclass[10pt, journal]{IEEEtran}

\usepackage[utf8]{inputenc}
\usepackage[T1]{fontenc}

\usepackage[cmex10]{amsmath}

\usepackage{algorithm}
\usepackage{algpseudocode}

\usepackage{placeins}

\usepackage{tikz}

\usepackage[flushleft]{threeparttable}

\usepackage{subfig}

\usepackage{siunitx}

\usepackage{url}

\usepackage{graphicx}
\graphicspath{{.}}
\DeclareGraphicsExtensions{.pdf}

\DeclareMathOperator{\sgn}{sgn}
\DeclareMathOperator{\almafa}{k1}

\begin{document}

\title{Calculating Ultra-Strong and Extended Solutions for Nine Men’s Morris, Morabaraba, and Lasker Morris}

\author{Gábor E. Gévay, and Gábor Danner \thanks{Gábor E. Gévay is with the Eötvös Loránd University, email: ggab90@gmail.com.} \thanks{Gábor Danner is with the University of Szeged, email: gabor.danner@gmail.com.}}

\maketitle

\begin{abstract}
The strong solutions of Nine Men’s Morris and its variant, Lasker Morris are well-known results (the starting positions are draws). We re-examined both of these games, and calculated extended strong solutions for them. By this we mean the game-theoretic values of all possible game states that could be reached from certain starting positions where the number of stones to be placed by the players is different from the standard rules. These were also calculated for a previously unsolved third variant, Morabaraba, with interesting results: most of the starting positions where the players can place an equal number of stones (including the standard starting position) are wins for the first player (as opposed to the above games, where these are usually draws). We also developed a multi-valued retrograde analysis, and used it as a basis for an algorithm for solving these games ultra-strongly. This means that when our program is playing against a fallible opponent, it has a greater chance of achieving a better result than the game-theoretic value, compared to randomly selecting between “just strongly” optimal moves. Previous attempts on ultra-strong solutions used local heuristics or learning during games, but we incorporated our algorithm into the retrograde analysis.

\end{abstract}

\section{Introduction} \label{sec:Intro}
Nine Men’s Morris and its variants are two-player, sequential, perfect information, deterministic, finite, zero-sum games, and there are three possible outcomes: win, draw, loss (these are given from the point of view of a specific player).

\subsection{Solving games}
Solving games is possible on several levels \cite{allisPhD}:
\begin{itemize}
\item Ultra-weakly solved: the game-theoretic value of the starting position was obtained by a possibly non-con\-struc\-tive proof, which does not give us any actual strategy to achieve the proven value.
\item Weakly solved: the game-theoretic value of the starting position is known, and we also have a strategy to achieve that. (This might require a database with the game-theoretic values for a large subset of the game states.) Checkers, for example, was solved in this sense \cite{schaeffer2007checkers}.
\item Strongly solved: a strategy is known that achieves the game-theoretic value starting from any game state that can be reached from the starting position. This has the effect that it can play perfectly even if mistakes were made on one or both sides.
\item Ultra-strongly solved \cite{schaeffer1996solving}: a strategy is known which increases our chances to achieve more than the game-theoretic value when faced with a fallible opponent (i.e.\ a player who is not playing perfectly).
\item Extended strong solution: We define this as a strong solution for an extended state space, namely, for all the positions reachable from a set of alternative starting positions. This can provide further insight into the game. In this paper, we examined the positions that can be obtained from the usual starting position by modifying the number of pieces to be placed by the players.
\end{itemize}

The standard method for strongly solving games is to use retrograde analysis (that is, to propagate values from end states using the minimax principle) to calculate a database containing the game-theoretic values of all the game states \cite{thompson1986retrograde}. (See Section \ref{sec:RA} for details.) A game-playing program can use this database at every move by looking ahead one move, and maximizing the game-theoretic value of the state (from its perspective) after its move. But if there are cycles in the state space graph, then this algorithm does not ensure winning. This might happen when we reach a game state during a game that has already occurred: if we choose our moves deterministically, then the game might never end, thus we cannot realize the game-theoretic value. Choosing randomly does not really solve this problem, because that way the games might stretch out too long.

The standard solution to this problem is to also calculate a \emph{depth to win} value for all (not draw) game states. This value gives the number of moves that will happen until the end of the game if both players play optimally, not just regarding the game-theoretic values, but also in minimizing or maximizing the number of moves to the end, based on whether they are winning or losing. This way, it cannot happen that we move away from the winning end state from time to time.

We used this method to calculate the extended strong solutions for standard Nine Men’s Morris, and its variants, Lasker Morris, and Morabaraba.

Strong solutions have an important property. Many of the positions among the theoretical draw positions are weak for one of the players for practical purposes (if he is not a perfect player), meaning that the player is just one small mistake away from getting into a theoretical losing position. In these cases the opponent is usually in an easier situation, because his small mistakes do not affect the game-theoretical value of the position. The program based only on the strong solution completely disregards this phenomenon, and tends to get to these weak positions. In the case of standard Nine Men’s Morris and Lasker Morris, this results in lots of draws, even against novice players. Gasser mentions this problem in his PhD dissertation \cite{gasserPhD}.

To solve this problem we developed a variant of retrograde analysis to classify the draws and used it to calculate ultra-strong solutions for all three variants.

\subsection{Rules}
For the rules of Nine Men’s Morris, see for example Gasser’s work \cite{gasser96}. There are two points in the rules, for which there is no consensus. The first question is what should happen when two mills are closed with one move. The second is whether a player should be allowed to take a stone from a mill, when all the opponent’s stones are in mills. Our implementation followed Gasser’s decisions: mill closure is always followed by taking exactly one stone. Moreover, we regard position repetitions as draws.

The main difference between Lasker Morris \cite{lasker1931brettspiele} and the standard variant is that there are no distinct placement and movement phases, i.e.\ the players can decide at every move whether they want to place a stone on the board or move one of their stones (as long as they have remaining stones to place). The other difference is that players can place 10 stones instead of 9.

Morabaraba differs from the standard variant in the graph of the game board (see Fig. \ref{fig:morastartpos} and Fig. \ref{fig:stdBoard}), and in the number of stones the players can place, which is 12. There is also a special rule: if the board becomes full (because neither player closed a mill during the opening), the game ends in a draw. There is no consensus about whether to use the rule or not, so we implemented both versions. The state space of Morabaraba is about four times bigger than that of the standard variant.

\subsection{Related work}
Ralph Gasser calculated a strong solution for the moving phase of the standard Nine Men’s Morris, and established the game-theoretic value of the game to be a draw \cite{gasser96}. One reason of the importance of Gasser’s work, is that he provided a perfect, and almost minimal hash function that takes the symmetries of the board into account, which allows an almost 16-fold reduction of the state space. Peter Stahlhacke calculated a strong solution for Lasker Morris \cite{Stahlhacke}. To our knowledge, there were no solution attempts for Morabaraba.

There are several approaches for achieving better results than the game-theoretic value against a fallible opponent. A simple method is to combine the perfect program with an AI that uses $\alpha$-$\beta$ search, by having the latter choose only from the perfect moves \cite{gasserPhD}. Other local heuristics use the information in the solution database about the game states which are at most a few moves away from the current position. For example, one can look at the ratio of optimal moves to all the moves in a position. The DTW (depth-to-win) of the non-draw states can also be relevant, because a model of a fallible opponent might assume that it is using shallow searches on the game-tree. These heuristics can be combined recursively by multiplying probabilities of making an optimal move along the considered paths \cite{NGM,Linckediploma}.
Schaeffer \cite[p.~258, 331]{schaeffer2008one} used small searches in positions recognized as draws by the endgame databases in his heuristic ($\alpha$-$\beta$) Checkers playing program Chinook.
Other approaches are to learn desirability values about individual positions during games \cite{ramsey}, or learn a model of the opponent \cite{opponentmodelling}.

Our approach is different, because we modified the retrograde analysis to calculate additional information about the game states, and developed a \emph{global}\footnote{Global in the sense that the evaluation of a game state calculated by our algorithm takes into account the heuristic values of all states reachable from that state.} heuristic to increase our chances of achieving more than the game-theoretic value.

{\v C}erm{\'a}k et al. studied \cite{cermak2014game} the performance of refinements of Nash Equilibria in smaller, imperfect-information games against fallible opponents. Their methods are computationally significantly more expensive, so they could not be used for Nine Men’s Morris.

A variation of retrograde analysis which can handle more than three outcomes was described by Lincke \cite{LinckePhD} for use with Awari, but that algorithm is quite different from ours. It uses less memory (but it is also slower), because it stores only two bits per positions, and does not calculate DTWs.


\subsection{The structure of the paper}
Although retrograde analysis is a well-known algorithm, implementations often differ in important details, because they are tailored to the actual problem. For example, an important point to note for Nine Men’s Morris and its variants is that there is a natural subdivision of the state space. In the next Sections we describe this partitioning, then briefly describe the basic retrograde analysis algorithm, which is followed by some implementation details along with a pseudocode.

Then we give a modified version of retrograde analysis which is able to handle more outcomes than win/draw/loss, and we use this as a basis for a new algorithm to classify draws into subclasses and achieve an ultra-strong solution. Finally, we present the results of the computations and then outline an extension and generalization of the algorithm.

\section{Partitioning the state space} \label{sec:partitioning}
Without partitioning the state space, retrograde analysis would require holding some information about every game state in memory, because random access disk I/O is very slow. A natural way to partition the state space of Nine Men’s Morris and its variants is to specify a subspace by four integers: the number of stones on the board for the first and second players, and the number of stones to be placed by the first and second players. The largest subspace has 603,332,730 game states (using Gasser’s hash function to take symmetries into account \cite{gasserPhD}).

Notice that if we swap the white and black stones and change the player to move in a particular position, then we get a game state which has the same game-theoretic value as the original (given from the point of view of the player to move). There are multiple ways to use this to achieve a 2-fold reduction of the state space. What we did is to drop every position where Black is to move. When we need the value of a position which was dropped, then we use the swapped position instead.

An unfortunate consequence of this is that the dependency graph\footnote{We can construct the following \emph{dependency graph} of the subspaces: each subspace corresponds to a node, and there is an arc from $u$ to $v$, if there is a state in $u$ from which we can go to a state in $v$ with one move.} of the subspaces ceases to be acyclic, since in the moving phase, we no longer have to take a stone from the board to leave a subspace. Rather, every sliding move in a subspace where the number of stones on the board (or the number of stones to be placed) for the two players are different, moves into the subspace where the first two and the last two identifying numbers are swapped (we call this operation \emph{negating a subspace}). So, for example, there are edges in both directions between the subspaces 5,7,0,0, and 7,5,0,0 (another example, which can occur only in Lasker Morris, is 5,7,2,1, and 7,5,1,2). The only cycles introduced this way are back and forth edges between subspaces, and a subspace can only be part of at most one cycle. We regard such subspace pairs (and also individual subspaces which are not part of a cycle) as \emph{work units}, because the retrograde analysis has to work with both of them at the same time. Let us call a work unit \emph{transient}, if every move in every position in it leaves the work unit. Notice that a transient work unit can contain only one subspace, which we call a \emph{transient subspace}. Let us call a subspace \emph{ESC} (Equal Stone Count) if both players have the same number of stones both on the board and to be placed.

Calculating the values of the positions in a work unit involves the one or two subspaces in the work unit (we call these \emph{primary subspaces}) and those other subspaces on which the primary subspaces directly depend, i.e.\ those for which there is a move from a position in a primary subspace that leads to a position in them (we call these \emph{secondary subspaces}). Information is propagated from secondary and primary subspaces to primary subspaces.

Another advantage of partitioning the state space is that pairs of work units which does not have a directed path between them in the dependency graph can be easily worked on in parallel. The scope of this is determined by the available memory (apart from the number of processor cores we have), because working on one of the larger work units requires multiple gigabytes of memory. The speed of the memory is also an important factor, since the frequency of memory accesses increases with more parallelism.

\section{Retrograde analysis} \label{sec:RA}

\subsection{The basic algorithm}

Retrograde analysis works from the ending positions backwards by calculating values according to the minimax principle \cite{strohlein1970untersuchungen}. To have an algorithm that can handle cycles in the state graph, we have to notice that in order to establish a position to be a win, we do not need the values of all the successors of the position. Rather, it is enough if we know that at least one of the successors is a loss (from now on, we are thinking as in negamax: position values are understood from the point of view of the player to move). To also correctly determine the DTW values when making use of the above point, we have to process positions in increasing order of DTW. Here, processing positions means repeatedly picking a position of which the final value is already known, and updating the knowledge we have about its predecessors.

Two kinds of information are kept for a position: \emph{count} means the number of successors that we have not processed yet, and \emph{value} means the game-theoretic value of the position (with DTW). The algorithm can be organized in such a way that these are not needed at the same time, and when we have a \emph{value} for a position, then it is already final. We say that a position is \emph{count-state} if the information currently recorded for it is a \emph{count}, and similarly for \emph{value-state}. If a position is still a count-state after no position remains to be processed, then it is a draw.

See figures \ref{fig:RA1} and \ref{fig:RA2} for the pseudocode.
\subsection{Implementation for a partitioned state space} \label{sec:RAImpl}

The question arises that how to pick positions to be processed. Gasser repeatedly scanned the entire file for positions that are done, and thus can be processed \cite{gasserPhD}. We decided to use a queue instead, for efficiency reasons. When a position becomes value-state, we push it into the queue. Picking a position to process is done by popping the queue. (This ensures the right order, because the DTW values always increase by one.)

There is some difficulty with this when the state space is partitioned: the range of DTW values can overlap between subspaces, so it would be inconvenient to process positions globally in the order of DTW. This means that sometimes we process a position that has a greater DTW value earlier than some positions that have lower DTW values. Notice that this is not a real problem: we only wanted to process the positions in increasing DTW order, to avoid the situation when processing some position affects our knowledge about some already processed positions (which would obviously create inconsistencies in the database). But if we keep the order locally, i.e.\ during the processing of one work unit, then this problematic situation does not occur. The reason is that the only case when we process a position $v$ with smaller DTW later than a position $u$ with larger DTW, is when the work unit of $v$ is processed later than the work unit of $u$, in which case the value of $v$ cannot possibly have an effect on the value of $u$, since a path in the state graph does not exist from $u$ to $v$.


\algrenewcommand\algorithmicindent{1.0em}%


\begin{figure}
\begin{algorithmic}[0]

\State 1.\ \emph{Initialization}
\ForAll{positions $p$}
	\If{$p$ is a win end-state}
		\State $R[p] \leftarrow$ \emph{value}(win in 0)
		\State Push $p$ into the priority queue
	\ElsIf{$p$ is a loss end-state}
		\State $R[p] \leftarrow$ \emph{value}(loss in 0)
		\State Push $p$ into the priority queue
	\ElsIf{$p$ is a draw end-state}
		\State $R[p] \leftarrow$ \emph{count}(0)
	\Else
		\State $R[p] \leftarrow$ \emph{count}(number of possible moves in $p$)
	\EndIf
\EndFor
\Statex

\State 2.\ \emph{Processing the priority queue}
\While{the priority queue is not empty}
	\State Pop a position $e$ from the priority queue
	\ForAll{predecessors $p$ of $e$}
		\If{$R[p]$ is \emph{count}}
			\If{$R[e]$ is a win}
				\State Decrement $R[p].count$
				\If{$R[p].count=0$}
					\State $R[p] \leftarrow$ \emph{value}(loss in $1+R[e].value.DTW$)
					\State Push $p$ into the priority queue
				\EndIf
			\Else
				\State $R[p] \leftarrow$ \emph{value}(win in $1+R[e].value.DTW$)
				\State Push $p$ into the priority queue
			\EndIf
		\EndIf
	\EndFor
\EndWhile

\end{algorithmic}
\caption{Pseudocode of retrograde analysis. Recall that the kinds of information recorded for a position (in the array $R$) are \emph{count}($n$) and \emph{value}(win/loss in $n$), \emph{but only one of these is stored at a time} (union type). (Also, in Nine Men’s Morris and its variants, win/loss can be determined by the parity of DTW.) The priority queue is keyed with DTW, and in the simplest case, it can be implemented with a simple queue, because states are processed monotonically. But when end states can have different DTWs (because we take into account the partitioning of the state space, see Fig. \ref{fig:RA2}), then the method of two queues can be used as described at the end of Subsection \ref{sec:RAImpl}. Note that states in secondary subspaces are treated as end states. Also note that $e$ can be in either a primary or a secondary subspace, but the predecessors are restricted to the primary subspaces.}\label{fig:RA1}
\end{figure}

\begin{figure}
\begin{algorithmic}[0]

\State 1. \emph{Pre-initialization of primary subspaces}
\ForAll{positions $e$ in primary subspaces}
\If{we can close a mill in $e$ and the opp. has 3 stones
}
		\State $R[e] \leftarrow$ \emph{value}(win in 1)
	\Else
		\State $R[e] \leftarrow$ \emph{count}(0)
	\EndIf
\EndFor
\Statex

\State 2. \emph{Setting \emph{count}s to the number of successors and pushing value-states}
\ForAll{positions $e$ in primary and secondary subspaces}
	\ForAll{predecessors $p$ of $e$}
		\If{$p$ is in a primary subspace and $R[p]$ is \emph{count}}
			\State Increment $R[p].count$
		\EndIf
	\EndFor
	\If{$R[e]$ is \emph{value}}
		\State Push $e$ into the priority queue
	\EndIf
\EndFor
\Statex

\State 3. \emph{Handling blocked states}
\ForAll{positions $e$ in primary subspaces}
	\If{$R[e]=$ \emph{count}(0)}
		\State $R[e] \leftarrow$ \emph{value}(loss in 0)
		\State Push $e$ into the priority queue
	\EndIf
\EndFor

\end{algorithmic}
\caption{Initialization of retrograde analysis tailored to Nine Men’s Morris and to the partitioning of the state space. This replaces Step 1 in Fig. \ref{fig:RA1}.} \label{fig:RA2}
\end{figure}

Thus we have to make sure that every one of the value-states in the secondary subspaces (which have a wide range of DTW values) gets processed at the right time, that is, when the processing of positions in the primary subspaces reaches the same DTW value. Theoretically thinking, this could be achieved by using a priority queue with DTW as the key instead of a regular queue, but a faster way in practice is to use two queues\footnote{Note that if we calculated the depths to the taking of the next piece instead of the DTW, then the problem would be trivially solved with only one queue, because then all the states in the secondary subspaces would come before all the states in the primary ones.} and continuously merge them when popping: one of the queues contains non-draw positions from the secondary subspaces (and end states from the primary ones) and is initialized at the start of processing the work unit, and the other is populated by positions from the primary subspaces as they become ready. We call the former the \emph{secondary queue}.

Initializing the secondary queue involves sorting the positions of several (in our case, up to six) subspaces, which might not fit into memory, so we used a bucket sort on disk to do this. The primary queue could be implemented on disk with only sequential accesses, but we stored it in main memory for simplicity and reducing disk I/O (for better parallelism).

The time complexity of the algorithm is $\mathcal{O}(E+VD+SB)$, where $V$ is the number of positions, $E$ is the number of edges between these, $D$ is the maximal in-degree of a node of the subspace graph, $S$ is the number of subspaces, and $B$ is the maximal DTW. The factor $B$ comes from the bucket sort. $VD$ comes from reprocessing positions multiple times as part of secondary subspaces. We assumed that the complexity of generating predecessors of a position is the number of predecessors that are in primary subspaces (without this assumption, the complexity would be $\mathcal{O}(ED+E+VD+SB)$).

\FloatBarrier

\subsection{Handling more than three outcomes} \label{sec:multivaluedRA}

We now modify the general algorithm described in the previous section to more than three outcomes. We assume that outcomes range from $-w$ to $w$. Our algorithm proceeds from more extreme outcomes to more drawish outcomes: we start by determining the positions that have values $-w$ and $w$, then $-w+1$ and $w-1$, and so on. While we are determining positions with absolute value $c$, the positions with absolute values smaller than $c$ have the same role as draws in the basic algorithm, and the positions with values $-c$ and $c$ will behave like losses and wins. The zeros are not processed.

At the beginning of the algorithm, non-zero end states are inserted into the secondary queue in the appropriate order (the first key of the ordering is the negated absolute value, and the second one is the DTW). Popping them at the right times can be achieved by the merging method we mentioned earlier.

In the basic algorithm, we made use of the fact that when one successor of a position $u$ becomes a loss, we know that $u$ is a win. There is a similar fact here, which points out the importance of going from larger absolute values towards smaller ones.

Assume that we are processing positions with absolute value $v$. If we discover a position with value $-v$ among the successors of a position $s$, then we immediately know that the value of $s$ is $v$. We will not find a successor $b$ later with value smaller than $-v$, because $b$ should have already been processed earlier. Furthermore, in the moment when we have processed all successors of a position $s$, then if none of them had negative values, then we know that the value of $s$ is $-v$ because so far we only processed nodes that have absolute values of at least $v$.

What we have showed in the previous paragraph is that when we write a value for a position, then it is indeed the correct final value of it. What remains to be proven is that we cannot have the situation that we should have already written a value for a particular position, but we have not. When we have finished writing the values of the positions with absolute values of $v$, then the absolute values of the positions that we have not yet written must really be $v-1$ at most, since it is not possible that all the successors of such a position has at least the value of $v$, or that it has a successor that has at most the value of $-v$.

The above algorithm also minimizes DTW values in “winning” (positive valued) states, and maximizes in “losing” (negative valued) states. Positions that remain count-states do not have DTW information. (Note that these can be viewed as implicitly having the value of 0.)

The pseudocode in the previous subsection can be adapted for this algorithm with small modifications:
\begin{itemize}
\item The comparison operator used by the queue operations has to use the negated absolute values of the states as a first key, and the DTW should only be the second key. Thus we process the positions with the same absolute values in the order of DTW.
\item Propagating a value to a predecessor now involves negating the game-theoretical value.
\end{itemize}

The time complexity changes a bit because of the extended range the bucket sort has to deal with: $\mathcal{O}(E+VD+SRB)$, where $R$ is the range of the first key.

This algorithm can be used for any game with the properties described at the beginning of the introduction, but with more than three possible outcomes, e.g.\ Awari or Othello.

\section{The ultra-strong solution}
The ultra-strong solution means that our program has a better chance to achieve a better result than the game-theoretic value against a non-perfect opponent, compared to a program which is based only on the strong solution. The outcome of a Nine Men’s Morris game (and its variants) can be win, draw, or loss. In (game-theoretically) winning states we are already in the best possible position (we are going to win, because we play perfectly). In losing states, maximizing DTW is already a good heuristic to lead the game into positions which are hard for the opponent to win. Therefore we designed an algorithm to classify draws into subclasses.

We need the notion of \emph{stable draw} for this: a draw is stable if there exists an optimal sequence of moves which does not leave the current work unit. (Not all draws are stable: positions in work units where the players have remaining stones to place are obviously not stable; furthermore, there are lots of positions in other work units where the only way for both players to keep the draw is to attack and close mills. Also notice that not all stable states are part of a cycle in the subgraph of optimal moves.)

Our method for distinguishing draws is based on the following goal: if we do not see a winning move, then at least try to go in the direction of stable draws where our opponent is in a (heuristically) difficult situation. Furthermore, it also seems reasonable to assume that when aiming for such draws, our opponent will “have a hard time” finding the moves that keep up the draw even before reaching the stable draw. Our results show that this is indeed the case: for example, in the standard variant, after our program has already managed to get to a game state from which optimal play leads to the subspace 6,3,0,0, our opponent almost always makes a mistake before even reaching the final work unit, and we win the game.

So we need a heuristic which can assign values to stable states, and then an algorithm which assigns different values to draws based on the value of the stable state that will be reached by optimal play. Notice that the algorithm cannot just pick the stable states at the beginning, use the heuristic to assign values to them, and then propagate these with standard multi-valued retrograde analysis, because we do not know in advance which states are stable, as this depends on the values of lots of other states. (Also notice that what was a stable state before we distinguished the draws is not necessarily stable now, so we also cannot just use these as a starting point.)

\subsection{The heuristic for assigning values to stable states} \label{sec:heur}

First, we describe the heuristic that is used to assign values to stable states. Actually, we assigned values to subspaces, and the value of a stable state is the same as the subspace in which it lies. A natural idea is to use the difference of the number of stones of the players.

We focused on another heuristic, where we used information from the already calculated databases: if the ratio of wins in a subspace is high, then it is obviously a good subspace for us (at least when the subspace is not ESC). The ratio of draws can also be relevant. If there is another subspace in the work unit, then a large ratio of losses in that subspace is also good for us (note that values in the other subspace are understood from the point of view of the opponent). Thus a pair of subspaces $(s,-s)$ in a non-transient work unit is in a symmetrical relationship (the members of the pair need not be different from each other). In Section \ref{sec:DD} it is shown that the algorithm works correctly only if the assigned values of these pairs are opposites of each other. So the heuristic formula we used takes both subspaces of the pair into account:
$$ val_{s} = (W_s + L_{-s} + D_s / 2 + D_{-s} / 2) / (T_s + T_{-s}) $$
where $W_s$, $L_s$, $D_s$ are the numbers of wins, losses, and draws, respectively, and  $T_{s}$ is the total number of states in the subspace. For subspaces in transient work units we use the following formula:
$$ val_s = (W_s + D_s / 2) / T_s $$

Using floating point numbers for subspace values during the computation would require too much memory, so we assigned ranks to subspaces instead: we ordered them based on the above values and used their place in the ordering. These ranks are centered around 0, and there is a correction to make: we have to make sure that the ranks of subspaces in a non-transient work unit are the negations of each other (for non-transient ESC subspaces, this means that they have to have the rank of 0). This correction results in holes in the range of ranks.

After using the ranking method described above, the rank of the subspace 8,9,0,0 seemed too high. For example, the program wanted to go to this subspace more than to 6,3,0,0. So we manually lowered the rank of this subspace, and the program got stronger (achieved more wins).

Wins and losses receive values just outside the range of the values calculated by the above ranking method. In the remaining part of the paper, game-theoretic values of states come from this extended range.

\subsection{A new variant of retrograde analysis to classify draws} \label{sec:DD}
This algorithm uses the heuristic of the previous section and is based on the multi-valued retrograde analysis described in Section \ref{sec:multivaluedRA}. Recall that the \emph{value} of a position now consists of two parts: the game-theoretic value and the DTW. We refer to these here as first and second keys, respectively. (Note that these keys are not directly the keys of an ordering.)

\subsubsection{Storing relative values} \label{sec:relativeValues}
The difficulty lies in the fact that with the algorithm described in Subsection \ref{sec:multivaluedRA}, the stable states can only be count-states (which can be viewed as implicitly having the value of 0 as the first key), but now we would need to assign different game-theoretic values to these. This is achieved by storing relative values in the first key: how much better (or worse) subspace than the current one we end up in with optimal play. This way, stable states look like having the appropriate first keys from the absolute viewpoint. For this, we need to adjust the first keys when propagating between subspaces: first, to the absolute viewpoint, then do the usual negation, then adjust to the value of the subspace that we are propagating to. Count-states are treated as if they had 0 as first key when making adjustments, and if the first key of a value-state is adjusted to 0, then it is treated as a count-state (the actual \emph{count} is not important in this case, since this can only happen in a secondary subspace, so we never propagate into it).

This can be implemented by already making all the adjustments while loading the secondary subspaces into memory (by adding the sum of the values of the two subspaces), because every state is propagated to only one primary subspace in Nine Men’s Morris\footnote{For other games, this can be achieved by the method described in Section \ref{sec:split}.}. The ordering of the queue has to be based on the values adjusted this way. Also notice that no adjustment is needed when propagating between primary subspaces, because the subspace values were constructed such that the sum of the values of the members of a non-transient work unit is 0.

\subsubsection{Generalizing depth-to-win}
It is not apparent how to extend the definition of DTW to draws, since it is not immediately clear that who should maximize or minimize it, and when. To determine this, we have to recall the intent of the players with minimizing and maximizing DTW. In the basic retrograde analysis, the winning player has to minimize DTW in order to avoid the situation where we are in a winning state, but cannot realize this, because we move away from it from time to time. When someone is losing against a perfect player, there is no point in maximizing DTW. Yet, we have to assume that he does that, because from the point of view of the winning player we have to “prepare for the worst”. In other words, for the minimization of DTW to work, we must do the opposite for the other player, because a minimax-based approach can only optimize for symmetric (zero-sum) utility functions.

Now the question is, should we be maximizing or minimizing when the state is a draw so there is no clear winning or losing player? As we mentioned earlier, we are seeking good valued stable states not just because being in the final subspace will be good for us, but we hope that the opponent will make a mistake on the way to reach it. Thus how much time we spend in good valued subspaces even before reaching the final subspace is of importance. \emph{We would like to use the DTW to optimize for spending as much time (before reaching a stable state) as possible in subspaces which are better than the final one (cf. DTW in losing states).} In order for this to make sense, we should optimize for spending as little time as possible in a subspace that has a worse value than the final one (and also to avoid the situation that we do not progress into a better subspace).

This is a multi-objective optimization task which we solve by linear scalarization: we maximize the (signed) difference between the number of moves taken in subspaces that are better than the final one and the number of moves taken in subspaces that are worse than the final one. (Note that this is symmetric (zero-sum) for the players.) (Also note that it might very well happen that the optimal sequence of moves leads through better and worse subspaces then the final destination.)

This can be implemented in the following way:
\begin{itemize}
\item we increment DTW upon propagation as usual;
\item when we are comparing two \emph{values} with equal first keys, we decide on minimizing or maximizing the second one (DTW) based on the sign of the first key (which reflects the relation between the values of the current and final subspaces) (minimize when negative);
\item when a first key changes sign during an adjustment (see Subsection \ref{sec:relativeValues}), we negate the second one. This way, \emph{the steps of the optimal path which are taken in oppositely optimized subspaces are reflected in DTW with opposite signs}.
\end{itemize}

Note that the first two items correspond to the way the algorithm in Subsection \ref{sec:multivaluedRA} manages DTWs. The third item is not needed in that algorithm, because the values stored there are not relative (cf. Subsection \ref{sec:relativeValues}).

Also note that this is equivalent with the following implementation: we always maximize DTW, but we increment or decrement it based on the sign of the first key (increment when negative) and we negate the DTW at every move\footnote{We can choose the order of the negation and incrementation/decrementation operations either way, but we have to make sure that the negation of the first key happens together with the negation of the DTW.}. This way DTW always equals the aforementioned utility function (while in the other implementation it is the negated utility function when the first key is positive).

Also note that the traditional DTW can be viewed as a special case of the above if we consider the values of the subspaces of non-terminal states to be 0. (End-states can be considered to be in a virtual subspace: there is a virtual \emph{lose} subspace and a virtual \emph{win} subspace.) In this case, one of the terms of the aforementioned utility function is always 0 (depending on the sign of the first key).

A limitation of the algorithm is that if the optimal path goes through an $s_2$ subspace which has the same $v$ value as the final $s_1$ subspace, then DTWs only take into account the part of the path before $s_2$. This happens because states in $s_2$ that have $v$ as first keys from an absolute point of view are considered count-states, so we lose the DTW information coming from secondary subspaces. Nevertheless, this does not seem to be a big problem.

The following equality holds for the DTWs calculated by the above algorithm:
$$ d(g) = \sgn (\almafa(g)) \sum\limits_{i=0}^{n-1} \sgn (\almafa_g(g_i)), $$
where $g_i$ are the positions of the optimal path starting from $g$, $n$ is the number of the first count-state or end state on the optimal path, $k1$ gives the relative first key of a state, and $k1_g$  negates this, if the player to move is not the same as in $g$.

If we do not assume that the optimal path is known, then the above formula can be written in the following form:
$$ d(g)=\sgn(\almafa(g))\min_{p\in P_g}\sum_{v\in p}\sgn(\almafa_g(v)), $$
where $P_g$ denotes the set of optimal paths (by the first key from the absolute viewpoint) up until the first count-state or end state, and $v$ is a state on a path.

The heuristic could be enhanced a little if we would not increment DTW by only 1, but by a number that is dependent on how much better subspace we are in, than the final one. But the values of subspaces are on an ordinal scale (mainly because of practical reasons), which does not fit well to this, so we did not implement it.
\section{Results}
\FloatBarrier

The computed databases for the ultra-strong solutions, and other resources\footnote{The programs that can compute and use the databases, and the heuristic $\alpha$-$\beta$ program that we used for testing are also available with their sources.} are available at our website: \verb+https://www.inf.u-szeged.hu/~danner/mills+.

Table \ref{table:stats} shows some statistics about the variants. Note that the numbers for Lasker Morris are different from what Stahlhacke gave \cite{Stahlhacke}, because he used slightly different rules.

Some of the databases were computed on an Intel Core i7-2630QM (2 GHz) machine with 16 GB of memory. Strongly solving Morabaraba took approximately 2 days, ultra-strongly solving the standard variant took one and a half days. Computing the extended solution of Lasker Morris took the most time, about 9 days on the same machine. Calculating the extended solution of Morabaraba took about 5 days on an AMD Phenom II X4 955 (3.2 GHz) machine with 20 GB memory. (The memories were larger than that of average computers to allow greater parallelism.) Note that in the case of Lasker Morris, not only the state space is larger, but the average number of possible moves in a position is also larger.

\begin{table}
\centering
\begin{tabular}{lcccc}
Variant & \#states (extended) & W/D/L \% & max. DTW \\
Standard & 27bn (284bn) & $53.6/22.3/24.1$ & 206 \\
Lasker Morris & 133bn (398bn) & $52.5/14.3/33.2$ & 214 \\
Morabaraba & 112bn (284bn) & $60.7/4.6/34.7$ & 124 \\
\end{tabular}
\caption{Number of states (also for extended solutions), win/draw/loss ratios, and maximal DTW for the three variants}
\label{table:stats}
\end{table}


\subsection{Morabaraba}

The game-theoretic value of Morabaraba is win in 49. This is in contrast with the other two variants, which are draws. Fig. \ref{fig:morastartpos} shows the values of all possible first moves of Morabaraba. Because of the board symmetries, actually only four of these moves are different. Two of them result in a draw, and the other two are wins in 49 and 69 moves.

\begin{figure}
 \centering
\resizebox{\columnwidth}{!}{
  \includegraphics{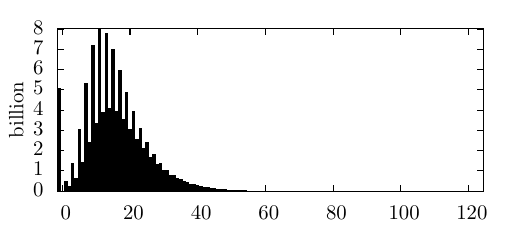}
 }
  \caption{Distribution of DTWs in Morabaraba. The very first bar shows the number of draws. Note that every second bar corresponds to an even depth, which means that it represents the number of \emph{losses} with that depth. These are usually smaller than neighbouring bars, because it is an advantage to have the right to move first.}
\label{fig:moradisthist}
\end{figure}

Fig. \ref{fig:moradisthist} shows the distribution of DTWs in Mo\-ra\-ba\-ra\-ba, and Fig. \ref{fig:moraoptimal} shows a game played optimally. Fig. \ref{fig:moramaxDTW} shows the position with the maximal DTW (124).

\begin{figure}
\centering
\resizebox{\columnwidth}{!}{
\begin{tikzpicture}[]
\tikzstyle{vertex}=[circle,draw,minimum size=10pt,inner sep=0pt]
\node[vertex] (S-0) at (0.567,-3.438) [label=180:\footnotesize $D$, minimum size=6pt, fill=black] {};
\node[vertex] (S-1) at (0.567,-0.576) [label=135:\footnotesize $W69$, minimum size=6pt, fill=black] {};
\node[vertex] (S-2) at (3.438,-0.576) [label=90:\footnotesize $D$, minimum size=6pt, fill=black] {};
\node[vertex] (S-3) at (6.309,-0.576) [label=45:\footnotesize $W69$, minimum size=6pt, fill=black] {};
\node[vertex] (S-4) at (6.309,-3.438) [label=0:\footnotesize $D$, minimum size=6pt, fill=black] {};
\node[vertex] (S-5) at (6.309,-6.309) [label=315:\footnotesize $W69$, minimum size=6pt, fill=black] {};
\node[vertex] (S-6) at (3.438,-6.309) [label=270:\footnotesize $D$, minimum size=6pt, fill=black] {};
\node[vertex] (S-7) at (0.567,-6.309) [label=225:\footnotesize $W69$, minimum size=6pt, fill=black] {};
\node[vertex] (S-8) at (1.431,-3.438) [label=135:\footnotesize $D$, minimum size=6pt, fill=black] {};
\node[vertex] (S-9) at (1.431,-1.431) [label=45:\footnotesize \bf{!}$W49$, minimum size=6pt, fill=black] {};
\node[vertex] (S-10) at (3.438,-1.431) [label=45:\footnotesize $D$, minimum size=6pt, fill=black] {};
\node[vertex] (S-11) at (5.445,-1.431) [label=135:\footnotesize \bf{!}$W49$, minimum size=6pt, fill=black] {};
\node[vertex] (S-12) at (5.445,-3.438) [label=315:\footnotesize $D$, minimum size=6pt, fill=black] {};
\node[vertex] (S-13) at (5.445,-5.445) [label=225:\footnotesize \bf{!}$W49$, minimum size=6pt, fill=black] {};
\node[vertex] (S-14) at (3.438,-5.445) [label=225:\footnotesize $D$, minimum size=6pt, fill=black] {};
\node[vertex] (S-15) at (1.431,-5.445) [label=315:\footnotesize \bf{!}$W49$, minimum size=6pt, fill=black] {};
\node[vertex] (S-16) at (2.295,-3.438) [label=0:\footnotesize $D$, minimum size=6pt, fill=black] {};
\node[vertex] (S-17) at (2.295,-2.286) [label=45:\footnotesize $W69$, minimum size=6pt, fill=black] {};
\node[vertex] (S-18) at (3.438,-2.286) [label=270:\footnotesize $D$, minimum size=6pt, fill=black] {};
\node[vertex] (S-19) at (4.59,-2.286) [label=135:\footnotesize $W69$, minimum size=6pt, fill=black] {};
\node[vertex] (S-20) at (4.59,-3.438) [label=180:\footnotesize $D$, minimum size=6pt, fill=black] {};
\node[vertex] (S-21) at (4.59,-4.581) [label=225:\footnotesize $W69$, minimum size=6pt, fill=black] {};
\node[vertex] (S-22) at (3.438,-4.581) [label=90:\footnotesize $D$, minimum size=6pt, fill=black] {};
\node[vertex] (S-23) at (2.295,-4.581) [label=315:\footnotesize $W69$, minimum size=6pt, fill=black] {};
\draw (S-0) -- (S-1);\draw (S-0) -- (S-7);\draw (S-0) -- (S-8);\draw (S-1) -- (S-0);\draw (S-1) -- (S-2);\draw (S-1) -- (S-9);\draw (S-2) -- (S-1);\draw (S-2) -- (S-3);\draw (S-2) -- (S-10);\draw (S-3) -- (S-2);\draw (S-3) -- (S-4);\draw (S-3) -- (S-11);\draw (S-4) -- (S-3);\draw (S-4) -- (S-5);\draw (S-4) -- (S-12);\draw (S-5) -- (S-4);\draw (S-5) -- (S-6);\draw (S-5) -- (S-13);\draw (S-6) -- (S-5);\draw (S-6) -- (S-7);\draw (S-6) -- (S-14);\draw (S-7) -- (S-0);\draw (S-7) -- (S-6);\draw (S-7) -- (S-15);\draw (S-8) -- (S-0);\draw (S-8) -- (S-9);\draw (S-8) -- (S-15);\draw (S-8) -- (S-16);\draw (S-9) -- (S-1);\draw (S-9) -- (S-8);\draw (S-9) -- (S-10);\draw (S-9) -- (S-17);\draw (S-10) -- (S-2);\draw (S-10) -- (S-9);\draw (S-10) -- (S-11);\draw (S-10) -- (S-18);\draw (S-11) -- (S-3);\draw (S-11) -- (S-10);\draw (S-11) -- (S-12);\draw (S-11) -- (S-19);\draw (S-12) -- (S-4);\draw (S-12) -- (S-11);\draw (S-12) -- (S-13);\draw (S-12) -- (S-20);\draw (S-13) -- (S-5);\draw (S-13) -- (S-12);\draw (S-13) -- (S-14);\draw (S-13) -- (S-21);\draw (S-14) -- (S-6);\draw (S-14) -- (S-13);\draw (S-14) -- (S-15);\draw (S-14) -- (S-22);\draw (S-15) -- (S-7);\draw (S-15) -- (S-8);\draw (S-15) -- (S-14);\draw (S-15) -- (S-23);\draw (S-16) -- (S-8);\draw (S-16) -- (S-17);\draw (S-16) -- (S-23);\draw (S-17) -- (S-9);\draw (S-17) -- (S-16);\draw (S-17) -- (S-18);\draw (S-18) -- (S-10);\draw (S-18) -- (S-17);\draw (S-18) -- (S-19);\draw (S-19) -- (S-11);\draw (S-19) -- (S-18);\draw (S-19) -- (S-20);\draw (S-20) -- (S-12);\draw (S-20) -- (S-19);\draw (S-20) -- (S-21);\draw (S-21) -- (S-13);\draw (S-21) -- (S-20);\draw (S-21) -- (S-22);\draw (S-22) -- (S-14);\draw (S-22) -- (S-21);\draw (S-22) -- (S-23);\draw (S-23) -- (S-15);\draw (S-23) -- (S-16);\draw (S-23) -- (S-22);
\end{tikzpicture}
}
\caption{Game-theoretic values with DTWs of all possible moves in the starting position of Morabaraba. The optimal moves are marked with an ``!’’.}
\label{fig:morastartpos}
\end{figure}

\begin{figure}
\centering
\resizebox{0.92\columnwidth}{!}{
\begin{tikzpicture}[]
 \tikzstyle{vertex}=[circle,draw,minimum size=10pt,inner sep=0pt]
  \node[vertex] (S-0) at (0.64,-3.82) [label=180:\footnotesize , minimum size=6pt, fill=black] {};
  \node[vertex] (S-1) at (0.64,-0.64) [label=135:\footnotesize , minimum size=13pt, line width=0.8pt] {};
  \node[vertex] (S-2) at (3.83,-0.64) [label=90:\footnotesize , minimum size=13pt, line width=0.8pt] {};
  \node[vertex] (S-3) at (7.02,-0.64) [label=45:\footnotesize , minimum size=13pt, line width=0.8pt] {};
  \node[vertex] (S-4) at (7.02,-3.82) [label=0:\footnotesize , minimum size=6pt, fill=black] {};
  \node[vertex] (S-5) at (7.02,-7.01) [label=315:\footnotesize , minimum size=13pt, line width=0.8pt] {};
  \node[vertex] (S-6) at (3.83,-7.01) [label=270:\footnotesize , minimum size=6pt, fill=black] {};
  \node[vertex] (S-7) at (0.64,-7.01) [label=225:\footnotesize , minimum size=13pt, line width=0.8pt] {};
  \node[vertex] (S-8) at (1.59,-3.82) [label=135:\footnotesize , minimum size=13pt, fill=black] {};
  \node[vertex] (S-9) at (1.59,-1.59) [label=45:\footnotesize , minimum size=13pt, fill=black] {};
  \node[vertex] (S-10) at (3.83,-1.59) [label=45:\footnotesize , minimum size=6pt, fill=black] {};
  \node[vertex] (S-11) at (6.05,-1.59) [label=135:\footnotesize , minimum size=6pt, fill=black] {};
  \node[vertex] (S-12) at (6.05,-3.82) [label=315:\footnotesize , minimum size=6pt, fill=black] {};
  \node[vertex] (S-13) at (6.05,-6.05) [label=225:\footnotesize , minimum size=13pt, line width=0.8pt] {};
  \node[vertex] (S-14) at (3.83,-6.05) [label=225:\footnotesize , minimum size=6pt, fill=black] {};
  \node[vertex] (S-15) at (1.59,-6.05) [label=315:\footnotesize , minimum size=6pt, fill=black] {};
  \node[vertex] (S-16) at (2.55,-3.82) [label=0:\footnotesize , minimum size=13pt, line width=0.8pt] {};
  \node[vertex] (S-17) at (2.55,-2.55) [label=45:\footnotesize , minimum size=6pt, fill=black] {};
  \node[vertex] (S-18) at (3.83,-2.55) [label=270:\footnotesize , minimum size=6pt, fill=black] {};
  \node[vertex] (S-19) at (5.1,-2.55) [label=135:\footnotesize , minimum size=13pt, line width=0.8pt] {};
  \node[vertex] (S-20) at (5.1,-3.82) [label=180:\footnotesize , minimum size=13pt, fill=black] {};
  \node[vertex] (S-21) at (5.1,-5.1) [label=225:\footnotesize , minimum size=6pt, fill=black] {};
  \node[vertex] (S-22) at (3.83,-5.1) [label=90:\footnotesize , minimum size=6pt, fill=black] {};
  \node[vertex] (S-23) at (2.55,-5.1) [label=315:\footnotesize , minimum size=6pt, fill=black] {};

  \node[vertex] (N-1) at (2.65,-3.52) [label=315:\footnotesize , minimum size=0pt, fill=black] {};
  \node[vertex] (N-2) at (2.65,-2.75) [label=315:\footnotesize , minimum size=0pt, fill=black] {};
  \draw [->] (N-1) -- (N-2);

\draw (S-0) -- (S-1);\draw (S-0) -- (S-7);\draw (S-0) -- (S-8);\draw (S-1) -- (S-0);\draw (S-1) -- (S-2);\draw (S-2) -- (S-1);\draw (S-2) -- (S-3);\draw (S-2) -- (S-10);\draw (S-3) -- (S-2);\draw (S-3) -- (S-4);\draw (S-4) -- (S-3);\draw (S-4) -- (S-5);\draw (S-4) -- (S-12);\draw (S-5) -- (S-4);\draw (S-5) -- (S-6);\draw (S-6) -- (S-5);\draw (S-6) -- (S-7);\draw (S-6) -- (S-14);\draw (S-7) -- (S-0);\draw (S-7) -- (S-6);\draw (S-8) -- (S-0);\draw (S-8) -- (S-9);\draw (S-8) -- (S-15);\draw (S-8) -- (S-16);\draw (S-9) -- (S-8);\draw (S-9) -- (S-10);\draw (S-10) -- (S-2);\draw (S-10) -- (S-9);\draw (S-10) -- (S-11);\draw (S-10) -- (S-18);\draw (S-11) -- (S-10);\draw (S-11) -- (S-12);\draw (S-12) -- (S-4);\draw (S-12) -- (S-11);\draw (S-12) -- (S-13);\draw (S-12) -- (S-20);\draw (S-13) -- (S-12);\draw (S-13) -- (S-14);\draw (S-14) -- (S-6);\draw (S-14) -- (S-13);\draw (S-14) -- (S-15);\draw (S-14) -- (S-22);\draw (S-15) -- (S-8);\draw (S-15) -- (S-14);\draw (S-16) -- (S-8);\draw (S-16) -- (S-17);\draw (S-16) -- (S-23);\draw (S-17) -- (S-16);\draw (S-17) -- (S-18);\draw (S-18) -- (S-10);\draw (S-18) -- (S-17);\draw (S-18) -- (S-19);\draw (S-19) -- (S-18);\draw (S-19) -- (S-20);\draw (S-20) -- (S-12);\draw (S-20) -- (S-19);\draw (S-20) -- (S-21);\draw (S-21) -- (S-20);\draw (S-21) -- (S-22);\draw (S-22) -- (S-14);\draw (S-22) -- (S-21);\draw (S-22) -- (S-23);\draw (S-23) -- (S-16);\draw (S-23) -- (S-22);
\end{tikzpicture}
}
\caption{A game-theoretical draw position in the standard or Lasker variant. White to move. Although White can preserve his material advantage, he cannot win. This is one of the only two such positions in the subspace of 8 white and 3 black stones. The only move here that can keep this stone count indefinitely is marked with an arrow.}
\label{fig:stdBoard}
\end{figure}

\begin{figure}
\footnotesize\raggedright\ttfamily{f2, f6; b6, b2; a1, c5; g1, e3; d1xf6, f6; g7, g4; a7, e4; e5, f4xe5; a4xe3, e3; e5, c3; d3, c4xd3; d3, d2; a7-d7, c4-b4; d7-a7xc5, f6-d6; g7-f6, g4-g7; g1-g4, c3-c4; a7-d7, c4-c3; g4-g1xg7, f4-g4; f6-f4, d6-d5; b6-a7xg4, d5-d6; e5-f6xd6, b4-c4; f6-g7xe3, c4-b4; g7-f6xe4, b4-b6; f6-g7xb2, c3-e5; f4-g4xd2}
\caption{A game of Morabaraba played optimally (using the standard notation for moves)}
\label{fig:moraoptimal}
\end{figure}

\begin{figure}
\centering
\resizebox{\columnwidth}{!}{
\begin{tikzpicture}[]
 \tikzstyle{vertex}=[circle,draw,minimum size=10pt,inner sep=0pt]
  \node[vertex] (S-0) at (0.63,-3.82) [label=180:\footnotesize -, minimum size=13pt, line width=0.8pt] {};
  \node[vertex] (S-1) at (0.63,-0.64) [label=135:\footnotesize $L38$, minimum size=6pt, fill=black] {};
  \node[vertex] (S-2) at (3.82,-0.64) [label=90:\footnotesize \bf{!}$L124$, minimum size=6pt, fill=black] {};
  \node[vertex] (S-3) at (7.01,-0.64) [label=45:\footnotesize $L32$, minimum size=6pt, fill=black] {};
  \node[vertex] (S-4) at (7.01,-3.82) [label=0:\footnotesize -, minimum size=13pt, fill=black] {};
  \node[vertex] (S-5) at (7.01,-7.01) [label=315:\footnotesize $L42$, minimum size=6pt, fill=black] {};
  \node[vertex] (S-6) at (3.82,-7.01) [label=270:\footnotesize $L36$, minimum size=6pt, fill=black] {};
  \node[vertex] (S-7) at (0.63,-7.01) [label=225:\footnotesize -, minimum size=13pt, line width=0.8pt] {};
  \node[vertex] (S-8) at (1.59,-3.82) [label=135:\footnotesize -, minimum size=13pt, line width=0.8pt] {};
  \node[vertex] (S-9) at (1.59,-1.59) [label=45:\footnotesize -, minimum size=13pt, line width=0.8pt] {};
  \node[vertex] (S-10) at (3.82,-1.59) [label=45:\footnotesize $L40$, minimum size=6pt, fill=black] {};
  \node[vertex] (S-11) at (6.05,-1.59) [label=135:\footnotesize $L42$, minimum size=6pt, fill=black] {};
  \node[vertex] (S-12) at (6.05,-3.82) [label=315:\footnotesize $L32$, minimum size=6pt, fill=black] {};
  \node[vertex] (S-13) at (6.05,-6.05) [label=225:\footnotesize -, minimum size=13pt, fill=black] {};
  \node[vertex] (S-14) at (3.82,-6.05) [label=225:\footnotesize -, minimum size=13pt, fill=black] {};
  \node[vertex] (S-15) at (1.59,-6.05) [label=315:\footnotesize -, minimum size=13pt, fill=black] {};
  \node[vertex] (S-16) at (2.55,-3.82) [label=0:\footnotesize -, minimum size=13pt, line width=0.8pt] {};
  \node[vertex] (S-17) at (2.55,-2.54) [label=45:\footnotesize $L24$, minimum size=6pt, fill=black] {};
  \node[vertex] (S-18) at (3.82,-2.54) [label=270:\footnotesize -, minimum size=13pt, fill=black] {};
  \node[vertex] (S-19) at (5.1,-2.54) [label=135:\footnotesize \bf{!}$L124$, minimum size=6pt, fill=black] {};
  \node[vertex] (S-20) at (5.1,-3.82) [label=180:\footnotesize $L36$, minimum size=6pt, fill=black] {};
  \node[vertex] (S-21) at (5.1,-5.09) [label=225:\footnotesize $L44$, minimum size=6pt, fill=black] {};
  \node[vertex] (S-22) at (3.82,-5.09) [label=90:\footnotesize $L30$, minimum size=6pt, fill=black] {};
  \node[vertex] (S-23) at (2.55,-5.09) [label=315:\footnotesize $L24$, minimum size=6pt, fill=black] {};
\draw (S-0) -- (S-1);\draw (S-0) -- (S-7);\draw (S-0) -- (S-8);\draw (S-1) -- (S-0);\draw (S-1) -- (S-2);\draw (S-1) -- (S-9);\draw (S-2) -- (S-1);\draw (S-2) -- (S-3);\draw (S-2) -- (S-10);\draw (S-3) -- (S-2);\draw (S-3) -- (S-4);\draw (S-3) -- (S-11);\draw (S-4) -- (S-3);\draw (S-4) -- (S-5);\draw (S-4) -- (S-12);\draw (S-5) -- (S-4);\draw (S-5) -- (S-6);\draw (S-5) -- (S-13);\draw (S-6) -- (S-5);\draw (S-6) -- (S-7);\draw (S-6) -- (S-14);\draw (S-7) -- (S-0);\draw (S-7) -- (S-6);\draw (S-7) -- (S-15);\draw (S-8) -- (S-0);\draw (S-8) -- (S-9);\draw (S-8) -- (S-15);\draw (S-8) -- (S-16);\draw (S-9) -- (S-1);\draw (S-9) -- (S-8);\draw (S-9) -- (S-10);\draw (S-9) -- (S-17);\draw (S-10) -- (S-2);\draw (S-10) -- (S-9);\draw (S-10) -- (S-11);\draw (S-10) -- (S-18);\draw (S-11) -- (S-3);\draw (S-11) -- (S-10);\draw (S-11) -- (S-12);\draw (S-11) -- (S-19);\draw (S-12) -- (S-4);\draw (S-12) -- (S-11);\draw (S-12) -- (S-13);\draw (S-12) -- (S-20);\draw (S-13) -- (S-5);\draw (S-13) -- (S-12);\draw (S-13) -- (S-14);\draw (S-13) -- (S-21);\draw (S-14) -- (S-6);\draw (S-14) -- (S-13);\draw (S-14) -- (S-15);\draw (S-14) -- (S-22);\draw (S-15) -- (S-7);\draw (S-15) -- (S-8);\draw (S-15) -- (S-14);\draw (S-15) -- (S-23);\draw (S-16) -- (S-8);\draw (S-16) -- (S-17);\draw (S-16) -- (S-23);\draw (S-17) -- (S-9);\draw (S-17) -- (S-16);\draw (S-17) -- (S-18);\draw (S-18) -- (S-10);\draw (S-18) -- (S-17);\draw (S-18) -- (S-19);\draw (S-19) -- (S-11);\draw (S-19) -- (S-18);\draw (S-19) -- (S-20);\draw (S-20) -- (S-12);\draw (S-20) -- (S-19);\draw (S-20) -- (S-21);\draw (S-21) -- (S-13);\draw (S-21) -- (S-20);\draw (S-21) -- (S-22);\draw (S-22) -- (S-14);\draw (S-22) -- (S-21);\draw (S-22) -- (S-23);\draw (S-23) -- (S-15);\draw (S-23) -- (S-16);\draw (S-23) -- (S-22);
\end{tikzpicture}
}
\caption{The Morabaraba position with the maximal DTW (124 plies). Both players have 7-7 pieces left to place.}\label{fig:moramaxDTW}
\end{figure}

\subsection{Extended solutions}
Tables \ref{table:stdfulltable}, \ref{table:laskerfulltable}, and \ref{table:morafulltable} show the game-theoretic values of the various starting positions involved in the extended solutions of the three variants. Uiterwijk and van den Herik investigated \cite{initiative} the advantage of the initiative (i.e.\ having the right to move first) in $mnk$-games and domineering on various board sizes. The extended solution tables for Nine Men’s Morris and its variants can also be examined in this respect.

The main diagonals divide the tables into positions where either White or Black has more stones to be placed. First, let us focus our attention to the cells not on the main diagonals. It is clear that having one more stone to be placed is a substantial advantage, since it completely outweighs the advantage of the initiative in all three variants: there are no wins for White if he can place fewer stones than Black, and these are even all losses above 5-6 stones to be placed.

However, if we also consider the DTWs in these cells, we can see a more fine-grained picture: almost all the wins for White below the main diagonals are quicker wins than the corresponding symmetrical positions for Black, so having the initiative means some advantage after all. (Note that this advantage is more pronounced in the non-standard variants.)

The diagonals show game-theoretic values of positions where the players can place an equal number of stones. These are all draws in Lasker Morris (except the 3-3 stones), and all wins in Morabaraba above 6-6 stones to be placed. However, in the standard variant, the values of the positions where the players can place 11 or 12 stones are losses for White!

We can conjecture about the reason for this surprising result. When we are placing a stone on the board, we can place it anywhere, but when we can only slide a stone, then our options are much more limited. This means that after Black places his last stone on the board, White can respond to the situation created by Black with (generally) less powerful moves. Also notice that after 22 (or 24) stones have been placed on the board, there is very little room left to slide pieces around (especially if one of the players played defensively during the opening), so there is a high chance that someone loses the game because of being unable to make a move. These two observations complement each other: Black has an advantage right when there is a high chance to end the game.

The role of the sometimes used rule that a full board results in a draw (rather than a loss for White) can also be examined. The only cells in the tables where it has a chance to make a difference are the bottom right cells. However, Morabaraba is a win for White, so this cannot possibly be affected by this rule. We found that this starting position is a draw with or without this rule in Lasker Morris, but in the standard variant this position changes into a draw from a loss upon introducing the rule.

There are a few places in the tables where one more stone for White results in a worse value. These are 3,3 and 11,12 for the standard variant, 3,3 for Lasker Morris, and 10,7 for Morabaraba. A similar place is 7,11 in Morabaraba where one more stone for Black results in a deeper loss for White.

\setlength{\tabcolsep}{4.1pt}

\begin{table}
\centering
\begin{tabular}{lllllllllll}
~ & 3 & 4 & 5 & 6 & 7 & 8 & 9 & 10 & 11 & 12 \\
3 & W23 & D & D & L12 & L12 & L12 & L12 & L12 & L12 & L12 \\
4 & D & D & D & D & L20 & L16 & L16 & L16 & L16 & L16 \\
5 & W9 & D & D & D & L24 & L22 & L20 & L20 & L20 & L20 \\
6 & W9 & D & D & D & L34 & L26 & L24 & L24 & L22 & L22 \\
7 & W9 & W17 & W25 & W33 & D & L36 & L28 & L26 & L26 & L24 \\
8 & W9 & W15 & W21 & W25 & W33 & D & L36 & L30 & L28 & L26 \\
9 & W9 & W15 & W19 & W23 & W27 & W35 & D & L36 & L32 & L28 \\
10 & W9 & W15 & W19 & W23 & W25 & W29 & W35 & D & L36 & L28 \\
11 & W9 & W15 & W19 & W21 & W25 & W27 & W29 & W31 & L44 & L28 \\
12 & W9 & W15 & W19 & W21 & W23 & W25 & W27 & W27 & W27 & *L26 \\
\end{tabular}
\caption{Game-theoretic values and DTWs of different starting positions of the standard Nine Men’s Morris. The first numbers of the rows and columns show the number of stones \emph{to be placed} by White and Black, respectively. The position marked with an ``*’’ becomes a draw, if we use the rule that a full board results in a draw. (This is the only case across all three variants where this rule makes a difference in the value of a starting position.)}
\label{table:stdfulltable}
\end{table}

\begin{table}
 \centering
    \begin{tabular}{lllllllllll}
    ~  & 3   & 4   & 5   & 6   & 7   & 8   & 9   & 10  & 11  & 12  \\
    3  & W23 & D   & D   & L12 & L12 & L12 & L12 & L12 & L12 & L12 \\
    4  & D   & D   & D   & L24 & L16 & L16 & L16 & L16 & L16 & L16 \\
    5  & W9  & D   & D   & D   & L24 & L20 & L20 & L20 & L20 & L20 \\
    6  & W9  & W19 & D   & D   & L40 & L26 & L24 & L22 & L22 & L22 \\
    7  & W9  & W15 & W21 & W39 & D   & L40 & L30 & L28 & L26 & L26 \\
    8  & W9  & W15 & W19 & W25 & W33 & D   & L44 & L34 & L32 & L30 \\
    9  & W9  & W15 & W17 & W23 & W29 & W35 & D   & L46 & L36 & L34 \\
    10 & W9  & W15 & W17 & W21 & W25 & W31 & W37 & D   & L44 & L38 \\
    11 & W9  & W15 & W17 & W21 & W25 & W29 & W33 & W39 & D   & L46 \\
    12 & W9  & W15 & W17 & W21 & W25 & W29 & W31 & W35 & W41 & D   \\
    \end{tabular}
\caption{Game-theoretic values and DTWs of different starting positions of Lasker Morris}
\label{table:laskerfulltable}
\end{table}

\begin{table}
 \centering
    \begin{tabular}{lllllllllll}
    ~  & 3  & 4   & 5   & 6   & 7   & 8   & 9   & 10  & 11  & 12  \\
    3  & W9 & D   & L12 & L12 & L12 & L12 & L12 & L12 & L12 & L12 \\
    4  & W9 & D   & D   & L20 & L18 & L16 & L16 & L16 & L16 & L16 \\
    5  & W9 & D   & D   & L22 & L20 & L20 & L18 & L18 & L18 & L18 \\
    6  & W9 & W17 & W23 & D   & L26 & L26 & L24 & L24 & L22 & L22 \\
    7  & W9 & W13 & W19 & W25 & W45 & L32 & L30 & L26 & L26 & L28 \\
    8  & W9 & W13 & W19 & W21 & W27 & W53 & L50 & L34 & L32 & L30 \\
    9  & W9 & W13 & W17 & W21 & W25 & W31 & W43 & L56 & L40 & L36 \\
    10 & W9 & W13 & W17 & W21 & W23 & W29 & W37 & W51 & L54 & L40 \\
    11 & W9 & W13 & W17 & W21 & W25 & W27 & W33 & W39 & W45 & L52 \\
    12 & W9 & W13 & W17 & W21 & W25 & W27 & W31 & W33 & W37 & W49 \\
    \end{tabular}
\caption{Game-theoretic values and DTWs of different starting positions of Morabaraba}
\label{table:morafulltable}
\end{table}

\setlength{\tabcolsep}{6pt}

\setlength{\tabcolsep}{3.7pt}
\begin{table*}[t]
\centering
\begin{tabular}{ccccccccccc}
~ & 3 & 4 & 5 & 6 & 7 & 8 & 9 & 10 & 11 & 12 \\
3 & 83 / 0\textsuperscript{+} / 17 & 13 / 87 / 0 & 0\textsuperscript{+} / 99 / 0\textsuperscript{+} & 0 / 97 / 3 & 0 / 53 / 47 & 0 / 9 / 91 & 0 / 0\textsuperscript{+} / 100 & 0 / 0 / 100 & 0 / 0 / 100 & 0 / 0 / 100 \\
4 & 10 / 90 / 0\textsuperscript{+} & 0\textsuperscript{+} / 100 / 0\textsuperscript{+} & 0\textsuperscript{+} / 100 / 0\textsuperscript{+} & 0\textsuperscript{+} / 94 / 6 & 0\textsuperscript{+} / 27 / 73 & 0\textsuperscript{+} / 7 / 93 & 0\textsuperscript{+} / 0\textsuperscript{+} / 100 & 0\textsuperscript{+} / 0 / 100 & 0\textsuperscript{+} / 0 / 100 & 0\textsuperscript{+} / 0 / 100 \\
5 & 23 / 77 / 0 & 0\textsuperscript{+} / 100 / 0\textsuperscript{+} & 0\textsuperscript{+} / 100 / 0\textsuperscript{+} & 0\textsuperscript{+} / 93 / 7 & 0\textsuperscript{+} / 40 / 60 & 0\textsuperscript{+} / 12 / 88 & 0\textsuperscript{+} / 2 / 98 & 0\textsuperscript{+} / 0\textsuperscript{+} / 100 & 0\textsuperscript{+} / 0\textsuperscript{+} / 100 & 0\textsuperscript{+} / 0\textsuperscript{+} / 100 \\
6 & 40 / 60 / 0 & 23 / 77 / 0\textsuperscript{+} & 24 / 76 / 0\textsuperscript{+} & 15 / 82 / 3 & 7 / 56 / 37 & 2 / 27 / 70 & 1 / 6 / 94 & 0\textsuperscript{+} / 1 / 99 & 0\textsuperscript{+} / 0\textsuperscript{+} / 100 & 0\textsuperscript{+} / 0\textsuperscript{+} / 100 \\
7 & 91 / 9 / 0 & 92 / 8 / 0\textsuperscript{+} & 88 / 12 / 0\textsuperscript{+} & 74 / 25 / 1 & 47 / 38 / 15 & 21 / 37 / 42 & 8 / 16 / 76 & 3 / 4 / 93 & 2 / 1 / 98 & 1 / 0\textsuperscript{+} / 99 \\
8 & 100 / 0\textsuperscript{+} / 0 & 99 / 1 / 0\textsuperscript{+} & 98 / 2 / 0\textsuperscript{+} & 94 / 6 / 0\textsuperscript{+} & 80 / 16 / 4 & 54 / 28 / 18 & 28 / 24 / 47 & 13 / 11 / 75 & 7 / 3 / 90 & 5 / 1 / 94 \\
9 & 100 / 0\textsuperscript{+} / 0 & 100 / 0\textsuperscript{+} / 0\textsuperscript{+} & 100 / 0\textsuperscript{+} / 0\textsuperscript{+} & 99 / 1 / 0\textsuperscript{+} & 96 / 3 / 1 & 83 / 11 / 7 & 59 / 19 / 22 & 36 / 15 / 49 & 23 / 7 / 70 & 18 / 3 / 80 \\
10 & 100 / 0 / 0 & 100 / 0 / 0\textsuperscript{+} & 100 / 0\textsuperscript{+} / 0\textsuperscript{+} & 100 / 0\textsuperscript{+} / 0\textsuperscript{+} & 99 / 0\textsuperscript{+} / 0\textsuperscript{+} & 95 / 3 / 2 & 82 / 8 / 10 & 62 / 11 / 27 & 46 / 7 / 47 & 39 / 2 / 59 \\
11 & 100 / 0 / 0 & 100 / 0 / 0\textsuperscript{+} & 100 / 0\textsuperscript{+} / 0\textsuperscript{+} & 100 / 0\textsuperscript{+} / 0\textsuperscript{+} & 100 / 0\textsuperscript{+} / 0\textsuperscript{+} & 98 / 1 / 1 & 92 / 3 / 6 & 78 / 4 / 17 & 64 / 3 / 33 & 53 / 0\textsuperscript{+} / 47 \\
12 & 100 / 0 / 0 & 100 / 0 / 0\textsuperscript{+} & 100 / 0\textsuperscript{+} / 0\textsuperscript{+} & 100 / 0\textsuperscript{+} / 0\textsuperscript{+} & 100 / 0\textsuperscript{+} / 0\textsuperscript{+} & 99 / 0\textsuperscript{+} / 1 & 94 / 1 / 5 & 83 / 1 / 16 & 64 / 0\textsuperscript{+} / 36 & 0 / 0 / 100 \\
\end{tabular}
\caption{Win/draw/loss percentages (rounded) in the (extended) standard variant for subspaces where no stones are left to be placed (White to move). The first numbers of the rows and columns show the number of stones \emph{on the board} for White and Black, respectively. This table is the same for Lasker Morris.}\label{fig:laskRatioTable}
\end{table*}
\setlength{\tabcolsep}{6pt}

\setlength{\tabcolsep}{3.7pt}
\begin{table*}[t]
\centering
\begin{tabular}{ccccccccccc}
~ & 3 & 4 & 5 & 6 & 7 & 8 & 9 & 10 & 11 & 12 \\
3 & 83 / 0 / 17 & 20 / 78 / 2 & 3 / 92 / 5 & 0\textsuperscript{+} / 60 / 40 & 0 / 17 / 83 & 0 / 0\textsuperscript{+} / 100 & 0 / 0 / 100 & 0 / 0 / 100 & 0 / 0 / 100 & 0 / 0 / 100 \\
4 & 21 / 78 / 1 & 6 / 93 / 1 & 1 / 95 / 4 & 0\textsuperscript{+} / 36 / 64 & 0 / 8 / 92 & 0 / 0\textsuperscript{+} / 100 & 0 / 0 / 100 & 0 / 0 / 100 & 0 / 0 / 100 & 0 / 0 / 100 \\
5 & 43 / 57 / 0\textsuperscript{+} & 12 / 88 / 0\textsuperscript{+} & 9 / 89 / 2 & 3 / 53 / 44 & 1 / 18 / 82 & 0\textsuperscript{+} / 2 / 98 & 0\textsuperscript{+} / 0\textsuperscript{+} / 100 & 0\textsuperscript{+} / 0\textsuperscript{+} / 100 & 0\textsuperscript{+} / 0\textsuperscript{+} / 100 & 0\textsuperscript{+} / 0 / 100 \\
6 & 88 / 12 / 0 & 90 / 10 / 0\textsuperscript{+} & 82 / 18 / 0\textsuperscript{+} & 52 / 33 / 15 & 19 / 35 / 46 & 3 / 9 / 88 & 0\textsuperscript{+} / 1 / 98 & 0\textsuperscript{+} / 0\textsuperscript{+} / 100 & 0\textsuperscript{+} / 0\textsuperscript{+} / 100 & 0\textsuperscript{+} / 0\textsuperscript{+} / 100 \\
7 & 99 / 1 / 0 & 99 / 1 / 0 & 98 / 2 / 0\textsuperscript{+} & 88 / 10 / 2 & 60 / 24 / 16 & 25 / 24 / 51 & 6 / 8 / 87 & 1 / 1 / 98 & 0\textsuperscript{+} / 0\textsuperscript{+} / 100 & 0\textsuperscript{+} / 0\textsuperscript{+} / 100 \\
8 & 100 / 0\textsuperscript{+} / 0 & 100 / 0\textsuperscript{+} / 0 & 100 / 0\textsuperscript{+} / 0\textsuperscript{+} & 99 / 1 / 0\textsuperscript{+} & 91 / 6 / 3 & 67 / 15 / 18 & 32 / 15 / 53 & 10 / 6 / 84 & 2 / 1 / 96 & 1 / 0\textsuperscript{+} / 99 \\
9 & 100 / 0 / 0 & 100 / 0 / 0 & 100 / 0\textsuperscript{+} / 0\textsuperscript{+} & 100 / 0\textsuperscript{+} / 0\textsuperscript{+} & 99 / 1 / 0\textsuperscript{+} & 92 / 4 / 4 & 70 / 10 / 21 & 39 / 9 / 52 & 17 / 4 / 79 & 9 / 1 / 90 \\
10 & 100 / 0 / 0 & 100 / 0 / 0 & 100 / 0\textsuperscript{+} / 0\textsuperscript{+} & 100 / 0\textsuperscript{+} / 0\textsuperscript{+} & 100 / 0\textsuperscript{+} / 0\textsuperscript{+} & 99 / 1 / 1 & 91 / 3 / 7 & 70 / 6 / 24 & 45 / 5 / 50 & 31 / 2 / 67 \\
11 & 100 / 0 / 0 & 100 / 0 / 0 & 100 / 0 / 0\textsuperscript{+} & 100 / 0\textsuperscript{+} / 0\textsuperscript{+} & 100 / 0\textsuperscript{+} / 0\textsuperscript{+} & 100 / 0\textsuperscript{+} / 0\textsuperscript{+} & 97 / 1 / 2 & 87 / 2 / 11 & 68 / 3 / 29 & 56 / 0\textsuperscript{+} / 44 \\
12 & 100 / 0 / 0 & 100 / 0 / 0 & 100 / 0 / 0 & 100 / 0\textsuperscript{+} / 0\textsuperscript{+} & 100 / 0\textsuperscript{+} / 0\textsuperscript{+} & 100 / 0\textsuperscript{+} / 0\textsuperscript{+} & 98 / 0\textsuperscript{+} / 1 & 91 / 1 / 8 & 75 / 0\textsuperscript{+} / 25 & 0 / 100 / 0 \\
\end{tabular}
\caption{Win/draw/loss percentages (rounded) in Morabaraba for subspaces where no stones are left to be placed (White to move). (Note that the bottom right entry depends on whether we use the rule that a full board results in a draw.)}\label{fig:moraRatioTable}
\end{table*}
\setlength{\tabcolsep}{6pt}

Table \ref{fig:laskRatioTable} shows the win/draw/loss percentages for the (extended) standard and Lasker Morris variants for subspaces where all stones have been placed. Table \ref{fig:moraRatioTable} shows the same for Morabaraba. A notable fact is that in Morabaraba there are no draws at all in the 3,3 subspace, as opposed to the standard and Lasker Morris variants where the ratio of the draws is 0.16\%.

\begin{table}
\centering
\begin{tabular}{ccccccccccc}
~ & 3 & 4 & 5 & 6 & 7 & 8 & 9 & 10 & 11 & 12 \\
3 & 26 & 33 & 31 & 6 & 30 & 34 & 30 & 16 & 14 & 12 \\
4 & 32 & 9 & 28 & 156 & 112 & 112 & 110 & 26 & 20 & 18 \\
5 & 3 & 29 & 57 & 162 & 160 & 160 & 114 & 114 & 54 & 34 \\
6 & 7 & 157 & 163 & 167 & 184 & 186 & 173 & 169 & 169 & 135 \\
7 & 31 & 111 & 159 & 185 & 181 & 204 & 202 & 180 & 152 & 134 \\
8 & 33 & 111 & 153 & 185 & 203 & 196 & 202 & 202 & 180 & 162 \\
9 & 25 & 103 & 113 & 172 & 201 & 201 & 191 & 191 & 186 & 180 \\
10 & 15 & 19 & 113 & 168 & 179 & 181 & 189 & 192 & 193 & 176 \\
11 & 13 & 17 & 33 & 134 & 151 & 179 & 185 & 185 & 185 & 148 \\
12 & 11 & 15 & 31 & 128 & 125 & 161 & 161 & 175 & 147 & 0 \\
\end{tabular}
\caption{Maximal depth-to-win values for the (extended) standard variant for subspaces where no stones are left to be placed (White to move). This table is the same for Lasker Morris.}\label{tab:stdMaxvalTable}
\end{table}

\begin{table}
\centering
\begin{tabular}{ccccccccccc}
~ & 3 & 4 & 5 & 6 & 7 & 8 & 9 & 10 & 11 & 12 \\
3 & 16 & 23 & 27 & 32 & 34 & 34 & 16 & 8 & 8 & 8 \\
4 & 22 & 19 & 34 & 46 & 42 & 38 & 20 & 12 & 12 & 10 \\
5 & 26 & 35 & 33 & 56 & 60 & 56 & 54 & 38 & 27 & 27 \\
6 & 33 & 45 & 57 & 81 & 86 & 86 & 78 & 56 & 48 & 39 \\
7 & 33 & 41 & 59 & 87 & 101 & 102 & 96 & 96 & 85 & 69 \\
8 & 33 & 31 & 55 & 83 & 101 & 103 & 102 & 106 & 108 & 94 \\
9 & 9 & 13 & 43 & 77 & 95 & 103 & 105 & 108 & 110 & 108 \\
10 & 7 & 11 & 25 & 53 & 93 & 107 & 109 & 111 & 112 & 114 \\
11 & 7 & 11 & 26 & 46 & 68 & 94 & 109 & 113 & 103 & 96 \\
12 & 7 & 9 & 13 & 34 & 51 & 93 & 106 & 106 & 90 & N/A \\
\end{tabular}
\caption{Maximal depth-to-win values for Morabaraba for subspaces where no stones are left to be placed (White to move)}\label{tab:moraMaxvalTable}
\end{table}

Tables \ref{tab:stdMaxvalTable} and \ref{tab:moraMaxvalTable} shows the maximal depth-to-win values in subspaces where all stones have been placed.


\subsection{Ultra-strong solutions}

We examined two heuristics for the values of stable states (see Subsection \ref{sec:heur}): one is based on stone difference and the other is based on win/draw/loss ratios in the work unit. In the following paragraphs we discuss the latter, except where noted otherwise.

Recall that in the databases for the ultra-strong solutions, the subspace which we end up in via optimal play determines the value of a draw. Table \ref{table:key1dist} shows the frequencies of these values (adjusted to absolute viewpoint) for the three variants. Unfortunately, 64.21\% and 55.66\% of the draw positions has the value of 0 in the standard variant and Lasker Morris, respectively. Table \ref{table:key1distStoneDiff} shows the same for the stone difference heuristic.

In the standard variant, the value of the starting position is also 0. This results in that lots of games still end with a draw (often in a 0-ranked subspace) even with our ultra-strong solution. However, the number of these games is substantially less than without distinguishing the draws.

The ultra-strong solution works better for Lasker Morris: here, the value of the starting position is 399. This means that when the program is the first player, it starts from a substantial advantage, and the opponent is just a small mistake away from losing the game.

To actually test the effectiveness of the ultra-strong solutions, we ran matches of the standard variant against a non-perfect program that uses $\alpha$-$\beta$ search\footnote{The program uses iterative deepening, transposition table, enhanced transposition cutoff, and killer move heuristic. It examines about 2-4 million nodes per second, and achieves $\sim$9-21 plies. The evaluation function takes into account the stone ratio, the number of available stone-sliding moves, and the possession of the points of the board that have four neighbors.}. The ultra-strong solutions performed significantly better than the strong solution. We also compared the two heuristics for assigning values to stable states. We expected the win/draw/loss ratio based heuristic to be better, but the results have not confirmed this. The win ratios can be seen in Table \ref{table:vsAlphaBeta}. In all our test games, the game was considered a draw if 50 consecutive moves happened without placing or taking a stone.

\setlength{\tabcolsep}{6pt}
\begin{table}
\centering
\begin{tabular}{lcccc}
~ & $0.1$-$0.2$ & $0.2$-$0.4$ & $0.4$-$0.8$ & $0.8$-$1.6$ \\
Ultra-str. (W/D/L rat. heur.) & $57\%$ & $42\%$ & $28\%$ & $25\%$ \\
Ultra-str. (stone diff. heur.) & $52\%$ & $51\%$ & $36\%$ & $32\%$ \\
Strong & $17\%$ & $21\%$ & $17\%$ & $12\%$ \\
\end{tabular}
\caption{Win ratios of the strong and the two ultra-strong solutions in the standard variant against the heuristic ($\alpha$-$\beta$) opponent with different node count settings: the numbers in the header give the number of nodes (in millions) searched by the heuristic program before each move (uniform random in the given interval, to provide greater variety to the games). Each match consisted of at least 200 games. (The sides were switched after each game.)}\label{table:vsAlphaBeta}
\end{table}
\setlength{\tabcolsep}{6pt}

We also did some testing at a website (\texttt{flyordie.com}) where usually humans play against each other.
With the standard variant, the ultra-strong program played 36 games against players titled ``master’’ in the website’s scoring system, and was able to win 18 of these (50\%). The strong program played 47 games against masters, and won only 11 (23\%). The ultra-strong program also played 19 games with the strongest player on the website (among thousands), of which our program was able to win 6 (32\%). Two of the games played against strong humans can be seen on Figs. \ref{fig:babi} and \ref{fig:ocsabi}. (The $\alpha$-$\beta$ program is just below master level on the same website.)

\begin{figure}
\footnotesize\raggedright{\ttfamily d7, d6; d2, f4; b6, b4; e4, c3; c4, d3; e3, e5; f2, d1; g7, a7?} (the value becomes the value of the 8,9,0,0 subspace; the only move that would have kept the value 0 is {\ttfamily c5}){\ttfamily;  b2xb4, b4; g7-g4, e5-d5; e4-e5, d1-a1; e3-e4, b4-a4xd7?} (now we are heading for the 3,6,0,0 subspace; the only move that would have kept the value is {\ttfamily a1-d1}){\ttfamily; c4-b4xd6, a1-d1; b4-c4, a4-b4; g4-g7, a7-d7; e4-e3, d5-d6?} (the value of the position after this move is win in 75 moves){\ttfamily; e5-d5, d7-a7; g7-d7, d1-a1; e3-e4, d3-e3} (win in 21 moves){\ttfamily; d2-d1, b4-a4xd7; c4-b4xf4, a7-d7; d1-d2xa1,} Black resigns
\caption{A won game of standard Nine Men’s Morris against a strong master player at the \emph{flyordie.com} website. Moves that do not keep the game-theoretic value (in the ultra-strong solution sense) are marked with a ``?’’. Our program played as White.}
\label{fig:babi}
\end{figure}

\begin{figure}
\footnotesize\raggedright{\ttfamily f4, b4; a4, d2; f6, f2; b2, d7; d1, g7; a7, a1; d5, g4; g1, c3; e3, c5; f6-d6, d2-d3; f4-e4, b4-c4xd5?; d6-d5, g4-f4; d5-e5xa1, c4-b4; e5-d5, b4-c4xd5; b2-b4, d3-d2; a4-a1xf4, g7-g4; e4-f4, c3-d3; b4-a4xc4,} Black resigns
\caption{A won game of standard Nine Men’s Morris against the strongest player at the \emph{flyordie.com} website. Our program played as White. The crucial mistake is made at the move marked with a ``?’’. The position after this move is a win in 41 moves. {\ttfamily b4-c4xe3} would have kept the draw (in a 0 valued subspace).}
\label{fig:ocsabi}
\end{figure}

In Lasker Morris, there were not enough strong players at the website to carry out proper testing, but the ultra-strong version seemed to be stronger here as well. In Lasker Morris, draws are much less frequent than in the standard variant (even with the strong solution). Recall, that according to our algorithm, the value of the starting position of Lasker Morris is quite high. The correctness of this is somewhat corroborated by the fact, that we won every one of the 16 games that the ultra-strong program played as the first player (often by the opponent making a fatal mistake in the first few moves), but there were two draws among the 17 games when playing as the second one.

We had the same problem with Morabaraba (not enough strong players). Additionally, when the ultra-strong program played Morabaraba as the second player, it usually either lost the game or the game transitioned directly from a losing position into a winning position. Unfortunately, this means that our ultra-strong solution did not help in these games, as only the draw-valued positions are treated differently from that of the strong solution. (In non-draw positions both optimize for DTW.)

We also experimented with some local heuristics to choose between the moves that the database says to be optimal, but these did not have a significant effect on the results. (For example, minimizing the number of optimal moves in the position that results from our move.)

\setlength{\tabcolsep}{4pt}
\begin{table}%
\centering
\hspace{-8pt}
\subfloat[][]{
\begin{tabular}{lrS[table-format=3.2]}
Subspace & Rank & \% \\
N/A & 0 & 64.21\% \\
6,3,0,0 & 112 & 15.6\% \\
3,6,0,0 & -112 & 12.34\% \\
3,5,0,0 & -60 & 1.3\% \\
5,3,0,0 & 60 & 1.19\% \\
5,4,0,0 & 29 & 0.78\% \\
9,8,0,0 & 59 & 0.72\% \\
8,7,0,0 & 113 & 0.65\% \\
7,8,0,0 & -113 & 0.54\% \\
8,9,0,0 & -59 & 0.53\% \\
Other & & 2.13\% \\
\end{tabular}
}%
\hspace{1pt}
\subfloat[][]{
\begin{tabular}{|lrS[table-format=3.2]}
Subspace & Rank & \% \\
N/A & 0 & 55.66\% \\
6,3,0,0 & 231 & 13.78\% \\
3,6,0,0 & -231 & 12.67\% \\
4,3,0,1 & -88 & 2.96\% \\
3,4,1,0 & 88 & 2.59\% \\
5,3,0,0 & 184 & 1.23\% \\
3,5,0,0 & -184 & 1.16\% \\
9,8,0,0 & 354 & 0.96\% \\
8,7,0,0 & 361 & 0.71\% \\
8,9,0,0 & -354 & 0.70\% \\
Other & & 7.58\% \\
\end{tabular}
}
\hspace{1pt}
\subfloat[][]{
\begin{tabular}{lrS[table-format=3.2]}
Subspace & Rank & \% \\
3,4,0,0 & -11 & 37.86\% \\
4,3,0,0 & 11 & 27.97\% \\
N/A & 0 & 11.51\% \\
5,3,0,0 & 54 & 8.38\% \\
3,5,0,0 & -54 & 7.69\% \\
5,4,0,0 & 28 & 1.01\% \\
4,5,0,0 & -28 & 0.96\% \\
11,10,0,0 & 86 & 0.71\% \\
10,11,0,0 & -86 & 0.65\% \\
10,9,0,0 & 99 & 0.40\% \\
Other & & 2.87\% \\
\end{tabular}
}
\caption{Distribution of first keys (adjusted to absolute viewpoint) among draws when draws are distinguished for standard Nine Men’s Morris (a), Lasker Morris (b), and Morabaraba (c)}%
\label{table:key1dist}
\end{table}
\setlength{\tabcolsep}{6pt}

\setlength{\tabcolsep}{4pt}
\begin{table}%
\centering
\hspace{-8pt}
\subfloat[][]{
\begin{tabular}{rS[table-format=3.2]}
Diff & \% \\
5 & 0\% \\
4 & 0\% \\
3 & 15.6\% \\
2 & 1.44\% \\
1 & 30.82\% \\
0 & 3.44\% \\
-1 & 34.95\% \\
-2 & 1.51\% \\
-3 & 12.23\% \\
-4 & 0.01\% \\
-5 & 0\% \\
\end{tabular}
}%
\hspace{1pt}
\subfloat[][]{
\begin{tabular}{|rS[table-format=3.2]}
Diff & \% \\
5 & 0.00\% \\
4 & 0.01\% \\
3 & 13.85\% \\
2 & 1.90\% \\
1 & 32.31\% \\
0 & 5.93\% \\
-1 & 31.62\% \\
-2 & 1.66\% \\
-3 & 12.73\% \\
-4 & 0.01\% \\
-5 & 0.00\% \\
\end{tabular}
}
\hspace{1pt}
\subfloat[][]{
\begin{tabular}{|rS[table-format=3.2]}
Diff & \% \\
5 & 0\% \\
4 & 0.01\% \\
3 & 0.07\% \\
2 & 8.88\% \\
1 & 30.73\% \\
0 & 11.51\% \\
-1 & 40.59\% \\
-2 & 8.14\% \\
-3 & 0.07\% \\
-4 & 0.01\% \\
-5 & 0\% \\
\end{tabular}
}
\caption{Distribution of first keys (adjusted to absolute viewpoint) among draws when draws are distinguished for standard Nine Men’s Morris (a), Lasker Morris (b), and Morabaraba (c) using the stone difference heuristic}
\label{table:key1distStoneDiff}
\end{table}
\setlength{\tabcolsep}{6pt}

\subsection{Verification of the calculations}
We took the approach of Gasser \cite{gasserPhD} for the verification of the databases: a separate verifier program went through all the positions, and checked if their values are consistent with the values of their successors (also taking into account DTWs).

\section{Conclusion}
We strongly solved Morabaraba, which turned out to be a win for the first player in 49 moves. We also calculated extended strong solutions for Nine Men’s Morris, Lasker Morris, and Morabaraba which provided some insights into these games (and also confirmed the result of Stahlhacke \cite{Stahlhacke} that Lasker Morris is a draw).

Furthermore, we developed a multi-valued retrograde analysis. Then we modified this to have an algorithm which can solve these games ultra-strongly. This means that the program has a much higher chance to win a game of Nine Men’s Morris or Lasker Morris against a fallible opponent instead of just drawing it, compared to a program which uses only a strong solution. This is important, because when a program is playing Nine Men’s Morris based only on the strong solution, it is surprisingly easy for the opponent to achieve a draw. The algorithm classifies draws into subclasses based on a heuristic value of the subspace that can be reached via perfect play. We investigated two heuristics for the subspace values.

We compared our ultra-strong solution to the strong solution by having them play against a heuristic ($\alpha$-$\beta$) program, and against human players, and found that it achieved wins more often.

\section{Future work}
\subsection{Splitting positions based on the type of the player to move}\label{sec:split}
When we are aiming for the strong solution, the players can be treated identically, but from the point of view of the ultra-strong solution, there is a perfect player, and a fallible one. (For simplicity, we refer to these as computer and human, respectively.) Note that previously we did not specify the type of the player to move in a given position. At the expense of doubling the state space, one can use more refined heuristics. Subspaces get a fifth parameter which specifies the player to move. This way, non-transient work units have exactly two subspaces: one human-to-move and one computer-to-move. Since these subspaces must have the opposite values, it is more convenient to talk about assigning values to work units. These values are the same as the values of the computer-to-move subspaces in the work units.

Now we can use a heuristic that assigns high value to a work unit where most of the computer-to-move positions are wins (and therefore the fallible player can easily wander into a loss), but disregards the values of the human-to-move positions (since the program does not make any mistakes when choosing the human-to-move position to move into). Previously we assigned neutral values to the subspaces of work units with many draws in it, but now we can assign low/high values to the computer/human-to-move subspaces of such work units. The motivation behind this is that a drawish work unit is bad for the perfect player if he wants to get his opponent to make a mistake, and is good for the human, since he can achieve a draw more easily (he cannot hope to win anyway).

Note that this might also solve the problem that too many subspaces had to be assigned the value of 0.

\subsection{Generalization to other games}

Our draw distinguishing algorithm can be used for any game with the properties described at the beginning of the introduction. But to be effective, there should exist a fine-grained enough partitioning of the state space, so that the partitions can be assigned meaningful heuristic values, and all cycles are confined to within one partition. In the general case, the splitting of positions to computer-to-move and human-to-move positions described in the previous subsection would be required\footnote{This way, it is also guaranteed that every position in a secondary subspace needs to be adjusted to only a single primary subspace.}. The partitions would correspond to the work units, and the subsets where a specific player type is to move would correspond to subspaces.

For example, a checkers playing program could probably benefit from this algorithm. The state space of that game is much larger, so our algorithm could only be used for some endgame databases, but that could still make a difference in some cases.

\section*{Acknowledgement}
We would like to thank the three anonymous reviewers (especially reviewer \#2) whose comments helped improve and clarify the paper.

\bibliographystyle{IEEEtran}
\bibliography{IEEEabrv,Games}

\end{document}